\documentclass[10pt,twocolumn,letterpaper]{article}

\usepackage{cvpr}              

\usepackage[dvipsnames]{xcolor}

\definecolor{cvprblue}{rgb}{0.21,0.49,0.74}
\usepackage[pagebackref,breaklinks,colorlinks,citecolor=cvprblue]{hyperref}

\def\paperID{8251} 
\def\confName{CVPR}
\def\confYear{2024}

\usepackage{amsmath,amsfonts}
\usepackage{algorithmic}
\usepackage{algorithm}
\usepackage{array}
\usepackage{textcomp}
\usepackage{stfloats}
\usepackage{url}
\usepackage{verbatim}
\usepackage{graphicx}
\usepackage{cite}
\usepackage{ulem} 
\usepackage{color}
\usepackage{setspace}
\usepackage{multirow}
\usepackage{tabularx}
\usepackage{booktabs}
\usepackage{ulem}
\usepackage{threeparttable}

\title{Context-Aware Integration of Language and Visual References for Natural Language Tracking}

\author{
{Yanyan Shao$^{1}$ \hspace{0.8cm} 
Shuting He$^{2}$ \hspace{0.8cm} 
Qi Ye $^{3}$\thanks{Corresponding author}  \hspace{0.8cm} 
Yuchao Feng$^{1}$\hspace{0.8cm} 
Wenhan Luo$^{4}$\hspace{0.8cm}  Jiming Chen$^{1,3}$ 
} \\
$^{1}$~Zhejiang University of Technology \hspace{0.3cm}   $^{2}$~Nanyang Technological University \hspace{0.3cm}   $^{3}$~Zhejiang University \\
$^{4}$~The Hong Kong University of Science and Technology
}

\begin{document}
\maketitle
\begin{abstract}

Tracking by natural language specification (TNL) aims to consistently localize a target in a video sequence given a linguistic description in the initial frame.
Existing methodologies perform language-based and template-based matching for target reasoning separately and merge the matching results from two sources, which suffer from tracking drift when language and visual templates miss-align with the dynamic target state and ambiguity in the later merging stage. To tackle the issues, we propose a joint multi-modal tracking framework with 1) a prompt modulation module to leverage the complementarity between temporal visual templates and language expressions, enabling precise and context-aware appearance and linguistic cues, and  2) a unified target decoding module to integrate the multi-modal reference cues and executes the integrated queries on the search image to predict the target location in an end-to-end manner directly.
This design ensures spatio-temporal consistency by leveraging historical visual information and introduces an integrated solution, generating predictions in a single step.
Extensive experiments conducted on TNL2K, OTB-Lang, LaSOT, and RefCOCOg validate the efficacy of our proposed approach. The results demonstrate competitive performance against state-of-the-art methods for both tracking and grounding. 
Code is available at \href{https://github.com/twotwo2/QueryNLT}{https://github.com/twotwo2/QueryNLT}

\end{abstract}
\section{Introduction}

\label{sec:intro}

\quad 
Tracking by natural language specification (TNL) aims to localize the target object in a video sequence based on a given language description on the initial frame. It offers a more user-friendly interaction to specify the target object compared to traditional tracking-by-bounding-box methods \cite{2019siamrpn++,guo2021graph,ye2022joint,cui2022joint},
which has a wide range of applications in video surveillance, robotics, and autonomous vehicles.

\begin{figure}[t]
  \centering
   \includegraphics[width=1 \linewidth]{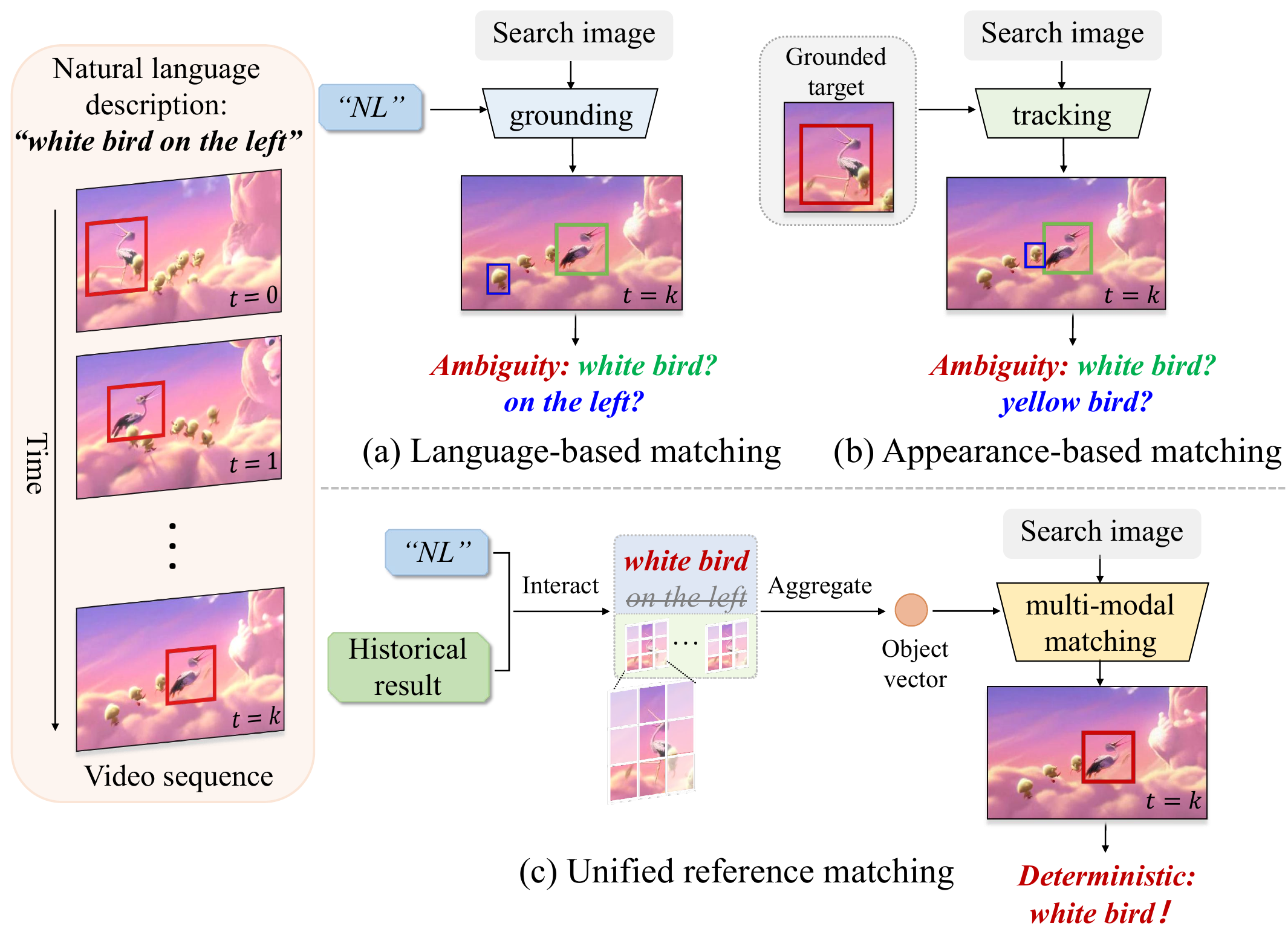}
 \caption{
   Given a video sequence, the tracking object is characterized as \textit{``white bird on the left''} of the initial frame. Existing two-step approaches separately perform language-search matching (a) and appearance-search matching (b).  However, \textit{``on the left"} which is inconsistent with the current target and the background contained in the grounded target may confuse the identification of the target. In contrast, our QueryNLT (c) forms a dynamic and context-aware query for target localization by integrating visual and language references. (Zoom in for a better view).
   }

   \label{fig:intro}
   \vspace{-10pt}
\end{figure}

Previous research efforts \cite{2017cvprLi,2020tcsvt_yang,2021_tnl2k_Wang,feng2019robust,2020wacv_feng} generally divide language-guided tracking into two fundamental sub-tasks: visual grounding and visual tracking. These studies initially localize the target object solely based on the given language description, i.e. visual grounding, and the grounded target serves as the visual template to establish correspondence with the search image, i.e. visual tracking. The final results are derived through the amalgamation of the outcomes obtained from both visual grounding and visual tracking.
Despite the significant success, these approaches typically process the language and template independently until merging their matching results, which may lead to ambiguity in target identification. As illustrated in Fig.~\ref{fig:intro}(a), due to the target's movement, the initially provided language description ({\textit{``white bird on the left"}) may no longer align with the current state of the target (the bird moves to the middle at $k^{th}$ frame).
This misalignment confuses the tracker's judgment of whether to focus on \textit{``white bird"} or \textit{``on the left"} during the matching process.
What is more, occlusions of objects may bring background clutters into the template. 
As shown in Fig.~\ref{fig:intro}(b), a small yellow bird is contained in the template, which may further interfere with the identification of the target ``\textit{white bird}".
The late fusion at the resultant level makes it difficult for the tracker to discern which candidate object is the real target, thus leading to tracking drift.

Based on the observation, we argue that the language description and the visual template are complementary and combining these two for matching contributes to a comprehensive understanding and perception of the target. 
To form the accuracy and context-aware target information as guidance, we propose a multi-modal prompt modulation module to filter out descriptions in the initial verbal reference and the visual reference accumulated in the tracking history that does not align with the current state. 
As illustrated in Fig.~\ref{fig:intro}(c), the historical results embed the target's motion information, which helps filter out status descriptions that fail to align with the actual target in the language expression. For instance, \textit{``on the left"} referring to a small yellow bird rather than the true target, should be removed.
Simultaneously, the categorization of the target, as depicted in the language description, serves as a reliable cue for filtering out extraneous background features in the visual template. Specifically, patches belonging to \textit{``white bird"} are assigned high attention weights, and patches belonging to the yellow bird and background are masked.
The revised language description collaborates with the accurate visual template to help point to the true target in the challenging scene.

Afterward, we present a query-based target decoding module that
jointly establishes the correspondence between the multi-modal references with the search image in a one-step fashion.
The key insight is to consider the language-based matching subtask and the appearance-based matching subtask as a unified instance-level retrieval problem.
To achieve this, this module comprises a multi-modal query generator that aggregates visual and verbal cues into a holistic object vector, and a query-based target locator that establishes the correspondence between the query vector and the search image for target retrieval. 
Compared with the previous works that need post-processing for merging results, it can directly predict the target location in an end-to-end manner. 
The prompt modulation module along with the target decoder module forms a unified framework to utilize the verbal and visual reference for natural language tracking.
With such a design, our proposed framework not only effectively improves the target discrimination through integrated perception, but also ensures the spatio-temporal consistency by 
forming context-aware query information.

We validate the effectiveness of our proposed framework through comprehensive evaluations on three tracking benchmarks and a grounding benchmark, including TNL2K \cite{2021_tnl2k_Wang}, OTB-Lang \cite{2017cvprLi,wu2015object}, LaSOT \cite{lasot}, and RefCOCOg \cite{refcocog}. Without bells and whistles, our QuertNLT achieves competitive performance compared with state-of-the-art trackers. Our main contributions are as follows.

\begin{itemize}
\itemsep 0pt
    \item We propose a novel framework for the natural language tracking task, termed QueryNLT. This framework integrates diverse modal references for target modeling and matching, fostering a holistic understanding of the target and improving discrimination capabilities.

    \item We propose a prompt modulation module that explores the complementarity of multi-modal reference to eliminate the inconsistent descriptions in the reference, generating precise and context-aware cues for target retrieval.
    
    
    \item We conduct comprehensive experiments on three challenging natural language tracking datasets and a visual grounding dataset, validating the efficacy of our proposed framework. The results showcase its robust performance and suitability for diverse tracking scenarios.
\end{itemize}
\section{Related Work}
\label{sec:related work}

\quad In this work, we aim to improve the performance of language-guided tracking by joining heterogeneous visual and language references. In the following, we will discuss related work that explores the utilization of these two heterogeneous references in existing language-guided tracking approaches, as well as how language-assistant target tracking approaches to underscore the potential benefits of the multi-modal tracking approach.

\subsection{Language-guidance Object Tracking}
\quad The emerging field of tracking by natural language specification (TNL) has garnered significant attention in recent years. It presents a unique approach to precisely localizing target objects within video sequences based on corresponding language descriptions. As the pioneering work in this area, Li \textit{et al.} \cite{2017cvprLi} first define the task of tracking by natural language specification and demonstrate the feasibility of language description replacing bounding boxes to specify targets. 
Subsequently, Yang \textit{et al.} \cite{2020tcsvt_yang} and Feng \textit{et al.} \cite{2020wacv_feng} share the same solution that divides this task into two subtasks: a grounding task solely relying on language to find the target and a tracking task based on the grounding results as the template. To better utilize the semantic information of the target during the tracking phase, \cite{2020tcsvt_yang} simultaneously performs visual matching based on the history of grounded objects, as well as performs grounding based on the language query for each subsequent frame. Besides, they propose an integration module to combine the prediction results of both processes adaptively. With the help of the region proposal network, \cite{2020wacv_feng} follows the tracking-by-detection formulation, leveraging language to select the most suitable proposal as the target template for tracking. 

In order to accelerate research for TNL, Wang \textit{et al.} \cite{2021_tnl2k_Wang} release a new benchmark and propose an adaptive switch framework that performs global search with language reference or local matching with visual template reference. While all of these approaches have made great progress, however, the grounding module used to initialize the template and the subsequent tracking used for tracking are separate, and cannot be trained end to end. Recently, Zhou \textit{et al.} \cite{2023joint_zhou} introduce a joint framework to replace a separate framework aiming at linking the language and template reference. However, it overlooks that the language expression may be inconsistent with the current tracking scene, which may cause references to be ambiguous. In this paper, we present a novel and effective framework that takes into account both linguistic descriptions and visual template information to improve target discrimination, while utilizing the complementary nature of heterogeneous information to form more accurate target reference information.

\subsection{Language-assisted Object Tracking}
\quad 
Different from the language-guidance tracking approaches the target object is specified only by the language description of the first frame, the tracking object of the language-assisted approaches \cite{2020arxiv_STL, 2021cvpr_snlt, 2022nips_guo, 2023cvpr_cite, wang2018describe,zhao2023mind,shao2023intertracker} is specified by both box and language.
With language description as an auxiliary cue, these works often focus on transforming a traditional box-guided tracking approach into a multi-modal target tracking approach.

Some work has been done to improve the performance of the tracker, in terms of improving visual feature representation \cite{2022nips_guo, wang2018describe} and enhancing the matching associations with the search image \cite{2021cvpr_snlt, 2023cvpr_cite}.
On the one hand, Feng \textit{et al.} \cite{2021cvpr_snlt} propose to perform symmetrical language-based matching alongside template-based matching \cite{2018cvpr_siamprn,2019siamrpn++}, where the results of both branches are weighted to obtain the final result. On the other hand, Guo \textit{et al.} \cite{2022nips_guo} treat language as a selector to reweight visual features and enhance visual feature representation through neural architecture search technology \cite{2017NAS,real2019regularized}. In contrast to the aforementioned methods where the language description is provided by the user, Li \textit{et al.} \cite{2023cvpr_cite} propose to automatically generate the corresponding semantic descriptions based on the input template. Taking advantage of the text-image alignment capability of CLIP models \cite{2021_clip}, \cite{2023cvpr_cite} designs to select the corresponding semantic descriptions from predefined attributes can be used as complementary descriptions.
These methods demonstrate that verbal cues alongside visual cues significantly enhance the overall understanding of the target, thus improving target discrimination.

\section{Method}


\subsection{Overview}
\quad
Our goal is to consistently and accurately localize the target within a video sequence, which is specified by language description. 
The main observation of our work is that the dynamic visual cue and the language expression provide complementary information that enhances target perception and discrimination.
Diverging from previous methods \cite{2021_tnl2k_Wang,2017cvprLi,wang2018describe} that employ separate networks for language-based and template-based matching, our proposed QueryNLT treats these two sub-tasks as an instance retrieval problem. To this end, we propose a unified multi-modal matching network for language-guided tracking. 

The framework of QueryNLT is depicted in Fig.~\ref{fig:framework}.
During the tracking phase, we collect the appearance feature $h_{a}$ and positional feature $h_{p}$ of the target based on the historical localization results of the target and store in a template memory $\mathcal{M}=\{h_a, h_p\}$ as the dynamic visual reference.
Given a search image $I_s$ and a language description $\mathcal{D}$, we first employ a feature extraction module (in section~\ref{sec:backbone}) to obtain the search feature $f_s$ and language feature $f_l$, respectively. 
Subsequently, in section~\ref{sec: prompt_generation}, we utilize a prompt modulation module 
to filter out the inconsistent description in the initial verbal reference and the visual reference, thus forming more precise prompt information to guide the target location.
Finally, in section~\ref{sec:decoding}, a target decoding module is responsible for integrating the multi-modal prompts and performing target retrieval within the search image.

\begin{figure*}[t]
  \centering
   \includegraphics[width=0.8 \linewidth]{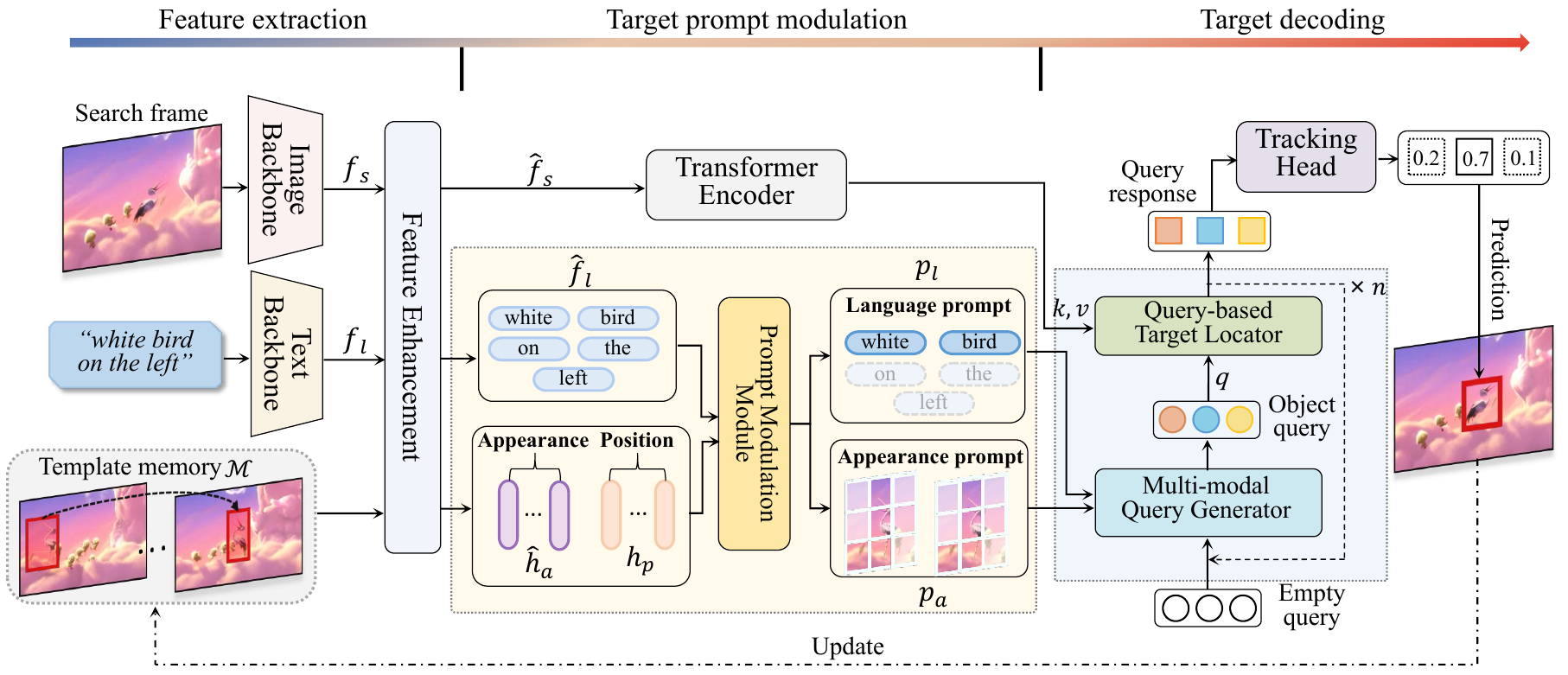}

   \caption{Overview of our proposed framework. It comprises three key components: a feature extraction module for extracting image and text features, a prompt modulation module that generates precise appearance and language descriptions of the target, and a target decoding module that
   jointly establishes the correlation between the search image and the multi-modal target prompts for target retrieval. }
   \vspace{-10pt}
   \label{fig:framework}
\end{figure*}

\subsection{Feature Extraction and Enhancement} \label{sec:backbone}
\quad \textbf{Visual backbone}.
Considering the notable achievements of transformer models in image processing, we adopt the vanilla Swin-Transformer \cite{liu2021swin} as our visual backbone. To strike a balance between tracking accuracy and computational cost, we retain only the first three stages of the Swin-Transformer architecture, with the output of the third stage serving as our visual feature representation.  
For the input search image $I_s \in \mathbb{R}^{3 \times H_s \times W_s}$, we feed it into the visual backbone and a channel adjustment layer 
to obtain the search region feature $\boldsymbol{f}_s \in \mathbb{R}^{N_s \times C}$,
where $N_s$ and $C$ denote the number of features and channels, respectively. Herein we set $C=256$.

\textbf{Text backbone}.
To process the language description $\mathcal{D}$, we employ the widely adopted linguistic embedding model, RoBERTa \cite{liu2019roberta}, for the extraction of textual features. A projection layer is added behind the text backbone for adjusting feature dimensions. The output text feature is denoted as $\boldsymbol{f}_l \in \mathbb{R}^{L \times C}$,  where $L$ represents the length of the input text and $C=256$ corresponds to the number of channels.
%

\textbf{Feature enhancement}.
To extract discriminative features, we employ a bi-attention mechanism between the search image and the target reference for feature enhancement.
In detail, the search feature $f_s$ attends to both the text feature $f_l$ and the historical appearance feature $h_a$ to obtain the enhanced search feature $\hat{f}_s$, while the text feature $f_l$ and the historical appearance feature $h_a$ separately attend to the search feature $f_s$ to obtain the enhanced feature $\hat{f}_l$ and $\hat{h}_a$. This process can be formulated as: 
{\setlength\abovedisplayskip{0.1cm}
\setlength\belowdisplayskip{0.1cm}
\begin{gather}
\label{eq_enhance}
\hat{f}_s = {\omega}_s(f_s + \mathrm{softmax}(\frac{f_s f_l^T}{\sqrt{C}}) f_l+ \mathrm{softmax}(\frac{f_s  h_a^T}{\sqrt{C}}) h_a),\\
\hat{f}_l = {\omega}_l(f_l + \mathrm{softmax}(\frac{f_l f_s^T}{\sqrt{C}})f_s), \\
\hat{h}_a = {\omega}_a(h_a + \mathrm{softmax}(\frac{h_a f_s^T}{\sqrt{C}})f_s),
\end{gather}
where ${\omega}_s$, ${\omega}_l$ and ${\omega}_a$ are linear layers. To avoid disturbing the motion information of the object, here we only augment the appearance feature in template memory.

\subsection{Target Prompt Modulation} \label{sec: prompt_generation}

\quad Accurate target cue information is essential for target tracking. However, due to the dynamics of the target in the course of tracking, the state description in language may not match the current target. 
As depicted in Fig.~\ref{fig:framework}, the positional state "on the left" in the language description corresponds to the object in the initial frame. However, as the object moves in subsequent frames, this description no longer aligns with the object. In fact, the object is currently in the middle of the image. 
Meanwhile, due to mutual occlusion between objects, the object's appearance features in the template memory may include background features. For instance, in Fig.~\ref{fig:framework}, the red bounding box erroneously encompasses the yellow bird, bringing further interference to the tracker. 
Using inaccurate language features and appearance features as the object prompt to retrieve the target in the search region may lead to tracking drift.
To address this issue, we present a multi-modal prompt modulation module that exploits the complementarity between dynamic historical information and the language description, facilitating the formation of a more accurate target prompt.

\textbf{Language modulation}.
We use motion cues from the template memory to adjust the language description. Specifically, $h_p$ in the $\mathcal{M}$ stores the object position information of multiple previous frames, serving as a motion cue to assess whether the state description in the text feature $\hat{f}_l$ aligns with the current scene. Meanwhile, $\hat{h}_a$ in the $\mathcal{M}$ contains the object appearance information, acting as a visual cue to evaluate whether the target appearance description in the text feature $\hat{f}_l$ is correct. 
As shown in Fig.~\ref{fig:modulation}(a), we utilize a multi-head cross-attention operation (MHCA) to generate the language prompt.
Before inputting to a cross-attention network, we first apply a self-attention operation on the $\hat{h}_a$ and $h_p$, respectively, to capture temporal changes in appearance and position. Subsequently, we add an appearance identifier vector $v_{a} \in \mathbb{R}^{C}$ to $\hat{h}_a$ and a motion identifier embedding $v_{p} \in \mathbb{R}^{C}$ to $h_p$. These two identifier vectors serve as indicators for different temporal cues. The process can be expressed by:
\begin{gather}
\hat{\mathcal{M}}= [\theta_a(\hat{h}_a) + [v_{a}]^{N_a}, \theta_p(h_p) + [v_{p}]^{N_p}],
\end{gather}
where $\theta_{a}$ and $\theta_{p}$ represent a self-attention operation. $[\cdot]^{n}$ denotes the duplicate the vector $n$ times and $[\cdot, \cdot]$ denotes the concatenation operation. ${N_a}$ and ${N_p}$ denotes the number of appearance feature and position feature saved in the template memory. The resulting $\hat{\mathcal{M}}$ is a matrix of size $N \times C$, where $N=N_a+N_p$. 

Next, we utilize $\hat{f}_l$ as Query and linearly transform $\hat{\mathcal{M}}$ to obtain Key and Value for cross attention. This process can be formulated as:
\begin{gather}
p_l = \varphi_{agg} (\hat{f}_l + \mathrm{softmax}(\frac{\varphi_q(\hat{f}_l) \varphi_k(\hat{\mathcal{M}})^T}{\sqrt{C}}) \varphi_v(\hat{m})),
\end{gather}
where $\varphi_{(\cdot)}$ represents different linear layer for feature transformation. 
After the above processing, the language features are re-weighted to generate the context-aware language prompt. 
Intuitively, the word that is more compatible with the template memory will be given higher attention, and the opposite will be given lower attention. We visualize the activation map of the language feature before and after modulation in Fig.~\ref{fig:vis_language}(b).

\begin{figure}[t]
  \centering
  \vspace{-10pt}
   \includegraphics[width=1. \linewidth]{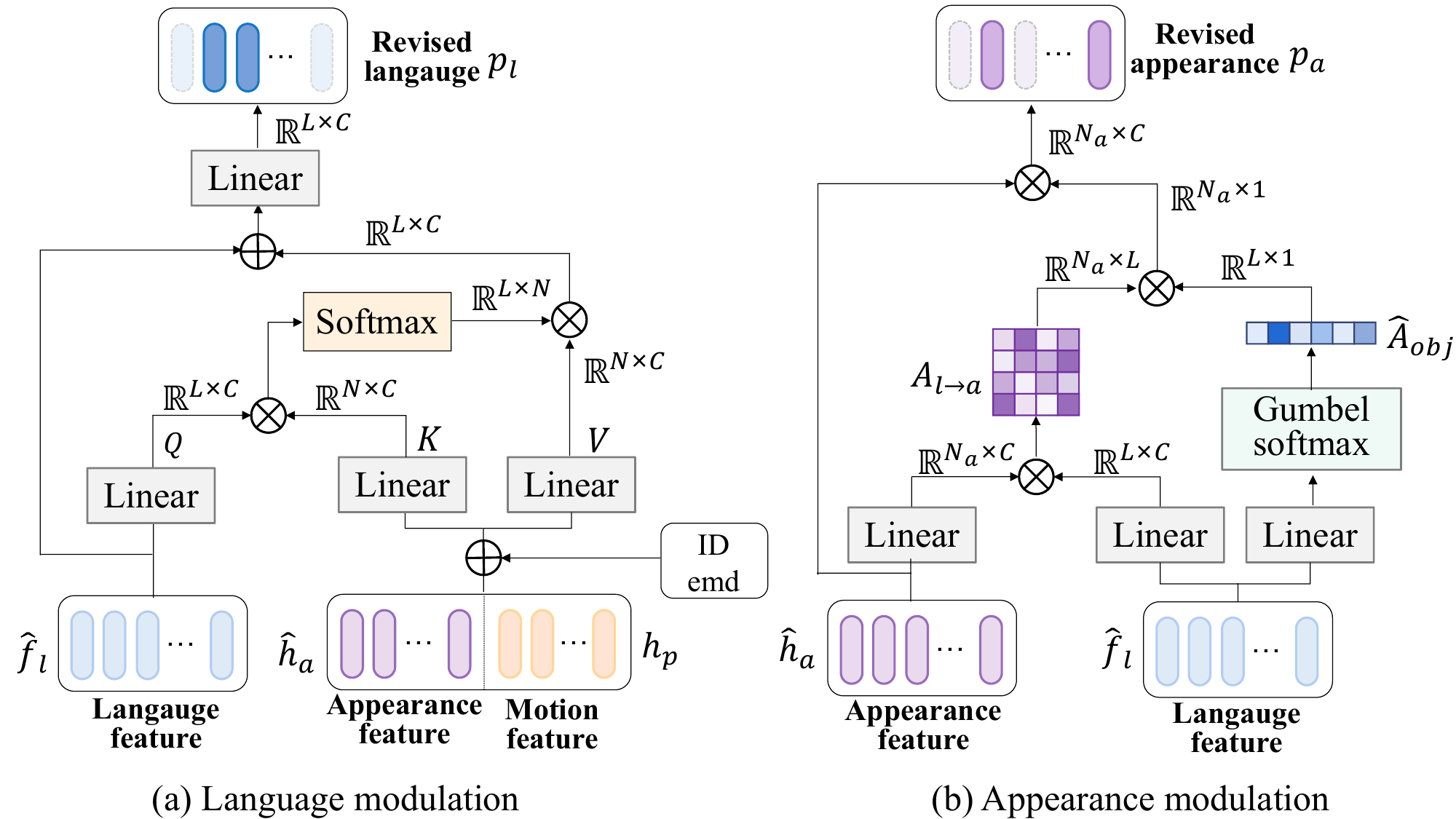}
   \caption{Architecture of the proposed language prompt modulation module (a) and the appearance modulation module (b).}
   \vspace{-15pt}
   \label{fig:modulation}
\end{figure}

\textbf{Appearance modulation}.
The purpose of appearance modulation is to generate a binarized mask based on the
category or appearance description of the tracked object in the sentence $D$,
which can better fit the shape of the target.
The appearance modulation is shown in Fig.~\ref{fig:modulation}(b). 
We first calculate a similarity matrix $A_{l \mapsto a}$ between $\hat{h}_a$ and $\hat{f_l}$:
\begin{gather}
A_{l \mapsto a} = \mathrm{softmax}(\frac{\delta_a(\hat{h}_a) \delta_l(\hat{f_l})^T}{\sqrt{C}}), 
\end{gather}
where $\delta_a$ and $\delta_l$ is a linear layer for feature transformation. This matrix $A_{l \mapsto a} \in \mathbb{R}^{N_a \times L}$ establishes the pixel-to-word correspondence, and pixels that correspond to the linguistic description yield high similarity scores.
However, the tracking objects are often specified by describing their relative position to other objects, such as \textit{``the fox on the bottom of the tree"}. In this case, the pixel belonging to the \textit{``tree"} also gets a high similarity score. 
Therefore, we compute binarized subjecthood scores $A_{obj}$ for the words via a Gumbel-Softmax \cite{jang2016categorical, maddison2016concrete} operation. For the $i^{th}$ word, its importance in the sentence is scored by:
\begin{gather}
A_{obj}^{i} = \frac{\mathrm{exp}(\mathbf{W}_{obj} \hat{f_l}^{i}+\gamma_i)}{\sum_{j=1}^{L} \mathrm{exp}(\mathbf{W}_{obj} \hat{f_l} + \gamma_j)}
\end{gather}
where $\mathbf{W}_{obj} \in \mathbb{R}^{1 \times C}$ is the weights of the learned linear projections for the text feature, $\gamma_i$ and $\gamma_j$ are random samples drawn from the Gumbel (0, 1) distribution. 
Then a subjecthood matrix score $\hat{A}_{obj} \in \mathbb{R}^{L \times 1}$ assigned all the words in the sentence is calculated by taking the one-hot \cite{xu2022groupvit} operation:
\begin{gather}
\hat{A}_{obj} = \mathrm{one-hot}(A_{obj}^{argmax}) + A_{obj}  - \mathrm{sg}(A_{obj}), 
\end{gather}
where $\mathrm{sg}$ is the stop gradient operator.
Finally, we multiply the $A_{l \mapsto a}$ and $\hat{A}_{obj}$ as the target mask, and the appearance prompt is formed by:
\begin{gather}
M_{obj} = A_{l \mapsto a} \times \hat{A}_{obj}, \\
p_a = \hat{h}_a \times M_{obj},
\end{gather}
where $M_{obj} \in \mathbb{R}^{N_a \times 1}$ indicates the probability that pixel belongs to the target. With such a design, we can filter out the background feature in the template, which produces a more accurate appearance prompt of the target for subsequent retrieval of the target.
\subsection{Target Decoding} \label{sec:decoding}
\quad 
Appearance information and language information are both important for accurate object tracking. 
Different from the previous works, we treat language-based matching and appearance-based matching as a unified instance tracking problem and propose a target decoding module to achieve it. 
This module is composed of a query generator that aims to produce a query vector derived from the language prompt and appearance prompt. A query-based target locator that establishes the correlation between the search image and the query vector.
The target decoding module is implemented by a transformer-based architecture proposed by a one-stage Deformable-DETR \cite{zhu2020deformable} for its flexible query-to-instance fashion.

As illustrated in Fig~\ref{fig:decoder}, an empty vector $q_{init}$ is concurrently injected into the prompt information through an attention-based mechanism. Then it transforms into an object-aware query vector $q_{obj}$. The process can be formulated as:
\begin{gather}
 q_{obj} = {\psi}(q_{init} + \alpha \sum_{i=0}^{N_t} a^i p_a^i + (1-\alpha)  \sum_{j=0}^{N_l} b^j p_l^j),
\end{gather}
where ${\psi}$ is a linear layer, and $\alpha$ is a coefficient balancing the information from the different modal prompts.
$a^i$ and $b^j$ denote the attention weights assigned to $i^{th}$ element in appearance prompt and $j^{th}$ element in the language prompt, respectively. Take the $a^i$ for example, it is calculated by:
\begin{gather}
 a^i = \frac{\mathrm{exp}(\mathbf{W}_{a}p_a^i \mathbf{W}_{q} q_{init})}{\sum_{k=0}^{N_t} \mathrm{exp}(\mathbf{W}_{a} p_a^k \mathbf{W}_{q}q_{init})}, 
\end{gather}
where $\mathbf{W}_{a} \in \mathbb{R}^{1 \times C}$, $\mathbf{W}_{q} \in \mathbb{R}^{1\times C}$ represent a transformation matrix.
When there is only language description as the target cue in the first frame of the sequence, we set $\alpha$ to $0$ and the rest of the stage to a learnable parameter. In this way, our decoder can use the same parameters for single-modal or multi-modal target tracking.
\begin{figure}[t]
  \centering
  \vspace{-10pt}
   \includegraphics[width=1.0 \linewidth]{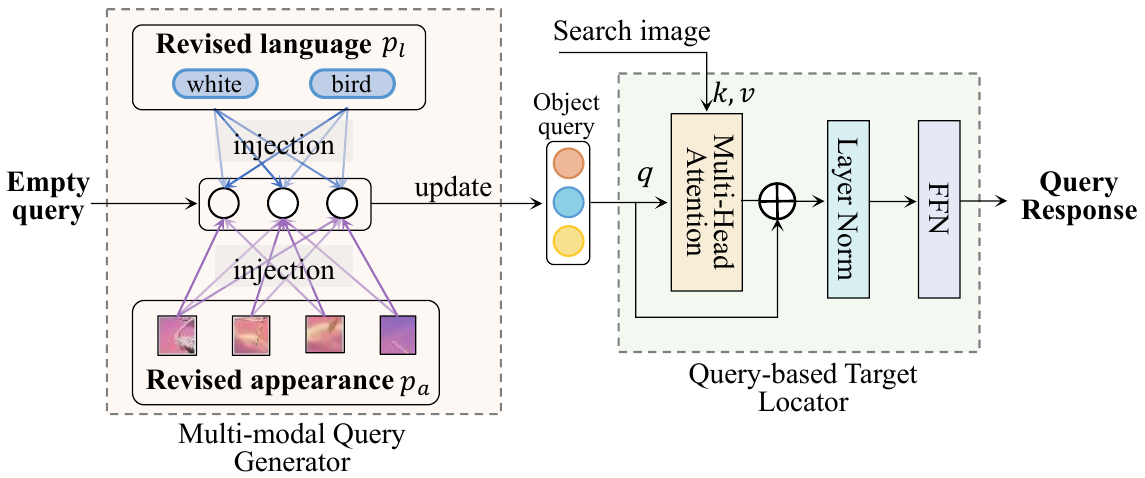}

   \caption{Architecture of the proposed target decoding module.}
   \vspace{-10pt}
   \label{fig:decoder}
\end{figure}
Afterward, in the query-based target locator, the object query $q_{obj}$ performed cross-attention with the search image to infer the target. In detail, 
the search feature $\hat{f}_s$ is fed into the transformer encoder network to serve as the Key and Value for the cross attention, with the query vector $ q_{obj}$ acting as the Query. 
The object query is iteratively refined over stacked decoder layers and outputs a query response $r$ which contains the target state within the search region.

To accommodate the diversity of linguistic expressions and target categories, which may result in a diversity of prompts, we adopt multiple vectors as a query set for target retrieval. Each query vector captures a unique interpretation of linguistic expressions and visual templates, emphasizing different aspects of the target. We conduct an ablation study on the number of queries as presented in Tab.~\ref{tab:ablation_query}.

Finally, we employ a classification head and a regression head to predict a score and box for each query. The bounding box with the highest score is selected as the final prediction. 
Following \cite{zhu2020deformable}, the regression head is supervised by the L1 loss and GIoU \cite{giou} loss, and the classification head is supervised by cross-entropy loss. 

After obtaining the position of the target in the current frame, we update the prediction result to template memory. In detail, the target feature is generated by a ROI align operation and is updated $h_a$. The corresponding query response which embeds the object's center position, width, and height in the current frame, is stored in $h_p$. Dynamically updated template memory facilitates the perception of the target's temporal changes.

\section{Experiment}
\label{sec:experiment}

\subsection{Implementation Details}
\quad The proposed QueryNLT is implemented in Pytorch on 6 NVIDIA RTX-3090 GPUs. We utilize Swin-B \cite{liu2021swin} pre-trained on ImageNet \cite{krizhevsky2017imagenet} as the visual backbone. 
The RoBERTa \cite{liu2019roberta} model is selected as the text backbone, with its parameters frozen throughout the entire training phase. The size of the template image and search image are set to $128 \times 128$ and $320 \times 320$, respectively. 
Following the training strategies of \cite{2023joint_zhou}, our training dataset comprises TNL2K \cite{2021_tnl2k_Wang}, OTB-Lang \cite{2017cvprLi}, LaSOT \cite{lasot} and RefCOCOg\cite{refcocog}, with an equal sampling ratio 
across the datasets.
The batch size is set to $12$ per GPU, with a total of $300$ epochs. 
We implement a warmup strategy where the initial learning rates for the visual and other parameters increase linearly to $10^{-5}$ and $10^{-4}$, respectively, within the first $30$ epochs. Subsequently, the learning rates are reduced by a factor of $10$ on the $200$-th and $290$-th epochs.



Following the protocols in \cite{2021_tnl2k_Wang}, we evaluate our approach with two settings: (1) \textit{``NL"}: the tracker is initialized with the natural language; (2) \textit{``NL+BB"}: the tracker is initialized with both the natural language and the bounding box. 
To ensure consistency in training and inference, we extract three frames from each video for training our network. During this process, the tracker is first initialized using linguistic descriptions. Subsequently, visual templates stored in memory are jointly utilized to predict the target in the subsequent frames. During the inference phase, We update the template memory by replacing outdated trajectories with new ones, limiting the memory to a maximum capacity of three frames.

\begin{table}  \small 
	\setlength{\abovecaptionskip}{0.cm}
	\setlength{\belowcaptionskip}{-0.4cm}	
	\vspace{-0pt}
        \caption{Ablation study on the components of QueryNLT. All models are trained on the same training set and evaluated on TNL2K \cite{2021_tnl2k_Wang} under the \textit{``NL"} setting.}
		\setlength{\tabcolsep}{2mm}
  	\begin{center}
		\begin{spacing}{0.8}
        \begin{tabular}{p{0.8cm}|p{5cm}|c}
        \toprule
            \#ID &  Model & $AUC$  \\
            \midrule                   
            0   & QueryNLT(Full Model)  &   53.3 \\
           \midrule
            1   & w/o language modulation &  52.2 \\
            2   & w/o appearance modulation &  51.6 \\
            3   & separate matching & 49.8 \\
            4   & static template  & 51.0 \\
        \bottomrule
        \end{tabular}
		\end{spacing}
	\end{center}
        \label{tab:ablation_component}
	\vspace{-10pt}
\end{table}

\begin{table}   \small 
	\setlength{\abovecaptionskip}{0.cm}
	\setlength{\belowcaptionskip}{-0.4cm}	
	\vspace{0pt}
        \caption{Ablation study on the query number on TNL2K \cite{2021_tnl2k_Wang} under the \textit{``NL"} setting. }
		\setlength{\tabcolsep}{2mm}
  	\begin{center}
		\begin{spacing}{0.8}
        \begin{tabular}{p{2.5cm}|cccc}
        \toprule
             Query number & 1 & 3 & 5 & 7 \\
            \midrule                   
             $AUC$   &  51.6 &  52.8 & 53.3 & 53.4 \\
        \bottomrule
        \end{tabular}
		\end{spacing}
	\end{center}
        \label{tab:ablation_query}
	\vspace{-20pt}
\end{table}

\begin{table*} \small
    \caption{Comparison of our method with state-of-the-art approaches on OTB-Lang \cite{wu2015object, 2017cvprLi}, LaSOT \cite{lasot} and TNL-2K \cite{2021_tnl2k_Wang} datasets. Top-2 results are highlighted in \textcolor{red}{red} and \textcolor{blue}{blue} respectively.}
    \vspace{-10pt}
    \begin{center}
    \begin{threeparttable}
    \setlength{\tabcolsep}{2mm}
    \begin{spacing}{0.9}
   \begin{tabular}{l|c|ccc|ccc|ccc}
    \toprule
         \multirow{2}*{Tracker}  &\multirow{2}*{Initialize}  &  \multicolumn{3}{c|}{OTB-Lang} & \multicolumn{3}{c|}{LaSOT} & \multicolumn{3}{c}{TNL-2K}   \\
        ~ &  ~ &  $AUC$ & $Prec$ & $NPrec$ & $AUC$ & $ Prec $ & $NPrec$ & $AUC$ & $ Prec $ & $NPrec$   \\
         \midrule
       SiamRPN++ \cite{2019siamrpn++} & BB & -  & -  & - & 49.6 & 49.1 & 56.9 & 41.3 & 41.2 & 48.0 \\ 
        Ocean \cite{zhang2020ocean}& BB & -  & -  & - & 56.0 & 56.6 & 65.1 & 38.4 & 37.7 & 45.0 \\ 
         AutoMatch \cite{zhang2021automatch}& BB & -  & -  & - & 58.3 & 59.9 & 67.4 & 47.2 & 43.5 & - \\ 
          TrDiMP \cite{TrDimp} & BB & -  & -  & - & 63.9 & 61.4 & - & 52.3 & 52.8 & - \\ 
        TransT \cite{TransT} & BB & -  & -  & - & 64.9 & 69.0 & 73.8 & 50.7 & 51.7 & - \\ 
        SwinTrack-B \cite{lin2022swintrack} & BB & -  & -  & - & 61.3 & 76.5 & - & - & 55.9 & 57.1 \\
        OSTrack-384 \cite{ostrack} & BB & -  & -  & - & 71.1 & 77.6 & 81.1 & 55.9 & - & - \\ 
           \midrule  
        TNLS-II \cite{2017cvprLi} & NL & 25.0  & 29.0  & - & - & - & - & - & - & - \\ 
        RVTNLN \cite{feng2019robust} &  NL & 54.0  & 56.0  & - & -  & -  & - & - & - & - \\ 
        
        RTTNLD \cite{2020wacv_feng} &  NL & 54.0  & \textcolor{blue}{78.0}  & - & 28.0  & 28.0  & - & - & - & - \\ 
        GTI \cite{2020tcsvt_yang} &  NL & 58.1  & 73.2  & - & 47.8  & 47.6  & - & - & - & - \\ 
        TNL2K-1 \cite{2021_tnl2k_Wang} &  NL & 19.0  & 24.0  & - & 51.1  & 49.3  & - & 11.4  & 6.4  & 11.0  \\ 
        CTRNLT \cite{li2022cross} & NL & 53.0  & 72.0  & - & 52.0  & 51.0  & - & 14.0  & 9.0  & - \\ 
        JointNLT \cite{2023joint_zhou} & NL & \textcolor{blue}{59.2}  & 77.6  & - & \textcolor{red}{56.9}  & \textcolor{red}{59.3}  & \textcolor{red}{64.5}  & \textcolor{red}{54.6}  & \textcolor{red}{55.0}  & \textcolor{red}{70.6}  \\ 
        JointNLT \cite{2023joint_zhou} \tnote{*} & NL & 57.8  & 77.0  & 70.5 & 52.8  & 54.4  & 60.8  & 52.1  & 51.2  &  68.8 \\ 
        Ours & NL & \textcolor{red}{61.2}  & \textcolor{red}{81.0}  & 73.9  &  \textcolor{blue}{54.2}  &  \textcolor{blue}{55.0}  &  \textcolor{blue}{62.5}  &  \textcolor{blue}{53.3}  &  \textcolor{blue}{53.0}  &  \textcolor{blue}{70.4}  \\ 
        \midrule
        TNLS-III \cite{2017cvprLi} & NL+BB & 55.0  & 72.0  & - & - & - & - & - & - & - \\ 

        RVTNLN \cite{feng2019robust} & NL+BB & 67.0  & 73.0  & - & 50.0  & 56.0  & - & 25.0 & 27.0 & 34.0 \\ 
                
        RTTNLD \cite{2020wacv_feng} &  NL+BB & 61.0  & 79.0  & - & 35.0  & 35.0  & - &  25.0 & 27.0  & 33.0 \\ 
        SNLT \cite{2021cvpr_snlt} &  NL+BB & 66.6  & 80.4  & - & 54.0  & 57.6  & - & 27.6  & 41.9  & - \\ 
        TNL2K-2 \cite{2021_tnl2k_Wang}  &  NL+BB & \textcolor{red}{68.0}  & \textcolor{blue}{88.0}  & - & 51.0  & 55.0  & - & 41.7  & 42.0  & 50.0  \\
        JointNLT \cite{2023joint_zhou} &  NL+BB & 65.3  & 85.6  & 79.5  & \textcolor{red}{60.4}  & \textcolor{red}{63.6}  & \textcolor{blue}{69.4}  & \textcolor{blue}{56.9}  & \textcolor{blue}{58.1}  & 73.6  \\ 
        JointNLT \cite{2023joint_zhou} \tnote{*} & NL+BB & 63.6  & 87.1  & 78.8 & 58.8  & 62.3  & 68.7  & 56.6  & 57.9 & \textcolor{blue}{74.8}   \\ 
        Ours &  NL+BB & \textcolor{blue}{66.7}  & \textcolor{red}{88.2}  & \textcolor{red}{82.4}  & \textcolor{blue}{59.9}  & \textcolor{blue}{63.5}  & \textcolor{red}{69.6}  & \textcolor{red}{57.8}  & \textcolor{red}{58.7}  &\textcolor{red}{75.6}  \\ 
        \bottomrule
    \end{tabular} 

	\begin{tablenotes}    
        \footnotesize               
        \item[*] our reproducing results using the officially released code.
    \end{tablenotes}
    \end{spacing}
    \vspace{-10pt}
    \end{threeparttable}
    \label{tab-trackingresults}
    \end{center}
\end{table*}

\subsection{Ablation Study}
\quad To assess the effectiveness of our proposed components, we conduct an ablation study on the TNL2K \cite{2021_tnl2k_Wang} dataset under the \textit{``NL"} evaluation setting. All variants are trained with the same training strategy as the full model. Tab.~\ref{tab:ablation_component} reveals the significance of each component. Line 2 and line 3 show that accurate target reference is important to improve the discrimination of the tracker. 
When replaced with the modulated language feature and modulated appearance, the AUC score improved from $52.2\%$ and $51.6\%$ to $53.3\%$, presenting improvements of $1.1\%$ and $1.7\%$, respectively.
Model 3 is a variant that disentangles the language-search matching and template-search matching within the proposed framework. The prediction with the highest score of the query set is considered the target prediction. 
When replaced with the full model that simultaneously establishes the correspondence between multi-modal reference with search image, there is a notable improvement of $2.5\%$ in AUC. This result demonstrates that the complementary nature of multi-modal information effectively boosts the holistic understanding and perception of the target. 
Model 4, relying solely on grounded results as the visual template, achieved an AUC score of $51.0\%$. However, introducing multiple dynamic templates led to a significant improvement from $51.0\%$ to $53.3\%$, underscoring the crucial role of temporal information in enhancing tracking robustness.

We provide an analysis of the impact of the query number for each frame as shown in Tab.~\ref{tab:ablation_query}. The model consistently demonstrates significant results across all settings. Overall, the performance improves with an increase in the query count. There is no noticeable improvement when the query number increases from 5 to 7. This observation suggests that a count of 5 already provides a sufficient variety of combinations to comprehend the cues. 
It is noteworthy that since queries are processed in parallel, an increase in the number of queries does not affect the speed of the tracker.

\subsection{State-of-the-art Comparison}
\quad In this section, we compare our approach with state-of-the-art trackers, including JointNLT \cite{2023joint_zhou}, CTRNLT \cite{li2022cross}, TNL2K \cite{2021_tnl2k_Wang} and others approaches, on three challenging natural language tracking datasets.
Following the protocols in \cite{2021_tnl2k_Wang}, we test the performance of our approach for initialization using only \textit{``NL''} and using \textit{``NL+BB''}. We also provide experimental results on the visual grounding dataset to demonstrate that our proposed method is effective in establishing text-image correlation.
All comparison results are obtained from the paper. Additionally, we retrained the JointNLT \cite{2023joint_zhou} using the official release code, denoted as ``JointNLT*''. By deploying in the same experimental setting, the result of  ``JointNLT*'' serves as an important baseline to measure the effectiveness of our method.

\begin{figure}[t]
\vspace{0pt}
  \centering
   \includegraphics[width=1 \linewidth]{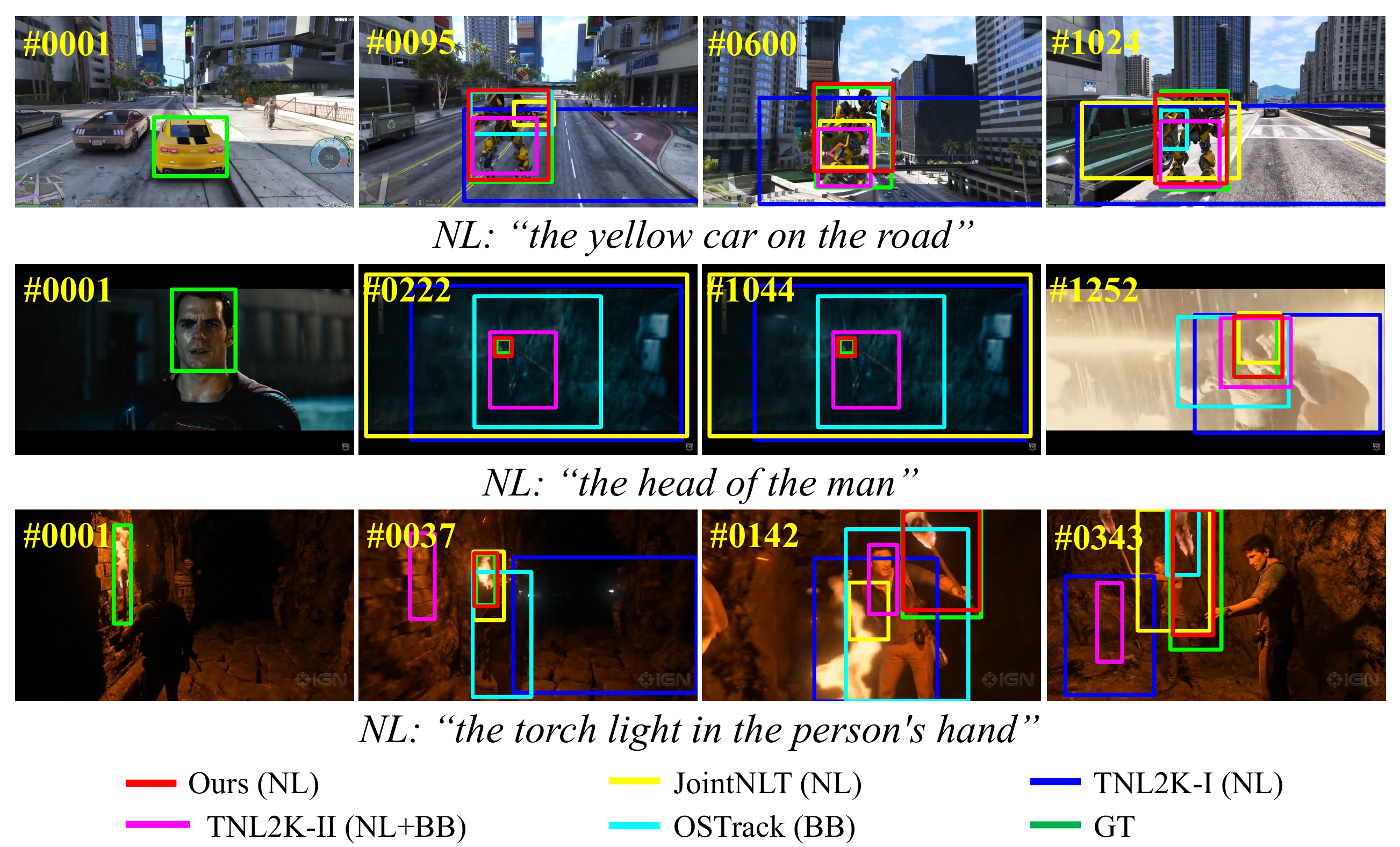}

   \caption{Qualitative comparisons of the proposed QueryNLT with the state-of-the-art trackers on three challenging sequences. Our QueryNLT can accurately target locations even when objects suffer from severe appearance variations, background clutters, and similar distractors.}
   \label{fig:comparsion}
   \vspace{-20pt}
\end{figure}

\begin{table}  \small 
    \caption{Comparison of our method with state-of-the-art approaches for visual grounding on RefCOCOg \cite{refcocog} dataset.}
    \vspace{-10pt}
    \begin{center}
    \setlength{\tabcolsep}{1mm}
    \begin{spacing}{0.9}
   \begin{tabular}{l|p{0.7cm}p{0.7cm}p{1.1cm}p{1.1cm}p{1.2cm}p{0.8cm}}
    \toprule
     Method &  \centering{LBYL} \cite{LBYL} & \centering ReSC \cite{ReSC-Large} & \centering TransVG \cite{deng2021transvg} & \centering VLTVG \cite{VLTVG} &  \centering{JointNLT} \cite{2023joint_zhou} &   \centering{Ours}  \tabularnewline
     \midrule
        val-g  &  \centering 62.70 &   \centering 63.12 & \centering 67.02 &  \centering \textcolor{red}{73.0} &  \centering 70.07 &  \centering \textcolor{blue}{72.0} \tabularnewline
        \centering val-u  & \centering - & \centering 67.3 & \centering 68.67 & \centering \textcolor{red}{76.0} & \centering - & \centering \textcolor{blue}{75.3} \tabularnewline
        \centering test-u  & \centering - & \centering 67.2 & \centering 67.73 & \centering \textcolor{red}{74.2} & \centering - & \centering \textcolor{blue}{73.2} \tabularnewline 
    \bottomrule
    \end{tabular} 
    \end{spacing}
    \end{center}
    \vspace{-20pt}
    \label{tab-groundingresults}
\end{table}

\textbf{Evaluation on TNL2K dataset}.
TNL2K is a benchmark specifically designed for evaluating natural language-guided tracking algorithms. 
It comprises a diverse collection of videos, including natural, animation, infrared, and virtual game videos, thereby facilitating a comprehensive evaluation of the framework's adaptability across different domains.
The rich and discriminative annotated language makes the TNL2K dataset particularly well-suited for the task of tracking based solely on natural language descriptions.
As shown in Tab.~\ref{tab-trackingresults}, 
under the \textit{``NL"}, our QueryNLT is the second best only behind the JointNLT \cite{2023joint_zhou} but surpasses the reproduction of the model JointNLT* by $1.2\%$, $1.8\%$ and $1.6\%$ on three metrics. 
Under the \textit{``NL+BB"} setting, our QueryNLT performs best in terms of all indicators. 
Compared with TNL2K-2 \cite{2021_tnl2k_Wang}, which employs adaptive switching between templates and language cues for target inference, our proposed approach (NL+BB) achieves notable improvements of $15.8\%$, $16.7\%$, and $25.1\%$ in terms of AUC, precision, and normalized precision, respectively. 
It demonstrates the complementary nature of multi-modal information in recognizing targets. Besides, our QueryNLT outperforms JointNLT \cite{2023joint_zhou}, which utilizes a static language description across all video frames, achieving superiority by $0.9\%$, $0.6\%$, and $2.0\%$ in terms of three metrics. 
The result emphasizes the effectiveness of dynamic and context-aware linguistic descriptions for improving tracking performance. Qualitative results are provided in Fig.~\ref{fig:comparsion}.

\textbf{Evaluation on OTB-Lang dataset}.
 The OTB-Lang dataset is originally released in \cite{wu2015object} and later extended with a sentence description of the target object per video by \cite{2017cvprLi}. 
 It encompasses $11$ challenging interference attributes, such as motion blur, scale variation, occlusion, out-of-view scenarios, background clutter, and more.
 The results on OTB-Lang are shown in Tab.~\ref{tab-trackingresults}. Remarkably, our proposed approach outperforms all other trackers under the \textit{``NL"} setting. Specifically, our proposed QueryNLT surpasses JointNLT \cite{2023joint_zhou} by $2.0\%$ and $3.4\%$ in terms of AUC and precision, respectively. And compared with JointNLT*, our approach shows improvements of $3.4\%$, $4.0\%$, and $3.4\%$ in three metrics. Additionally, under the \textit{``NL+BB"} setting, our QueryNLT is the second best in terms of AUC, only behind TNL2K \cite{2021_tnl2k_Wang}, within which the tracking module is trained on a larger training dataset. These results collectively highlight the robustness of our proposed approach, indicating its ability to effectively handle various challenging factors encountered in tracking tasks.

\textbf{Evaluation on LaSOT dataset}.
The LaSOT is a long-term tracking dataset that provides both bounding box and natural language annotations. It comprises $1120$ training video sequences and $280$ testing video sequences. 
It should be noted that the linguistic information in LaSOT lacks a description of the relative positions of the objects, and thus the given linguistic description is ambiguous when similar objects are interfering. This means that this dataset is not suitable for accomplishing language-assist tracking tasks, and a similar view can be found in \cite{2020tcsvt_yang, 2021_tnl2k_Wang}.
Here we mainly discuss the comparison results under the \textit{``NL+BB"} setting.
As shown in Tab.~\ref{tab-trackingresults}, our proposed approach 
achieves the performance of $57.8\%$  $58.7\%$, and $75.6\%$ in terms of AUC, precision, and normalization precision, respectively.
It surpasses the TNL2K-2 \cite{2021_tnl2k_Wang} by $5.9\%$ in AUC and $5.9\%$ in precision.
These results demonstrate that our approach is competitive for long-term tracking tasks.

\textbf{Evaluation on RefCOCOg dataset}.
We evaluate the visual grounding performance on both the validation and test sets of the RefCOCOg dataset \cite{refcocog}. The assessment is conducted using the average IoU as the evaluation metric."
As shown in Tab.~\ref{tab-groundingresults}, our method is second only to VLTVG \cite{VLTVG} although we are not specialized for visual grounding. 
This also explains that our tracking method can perform robust tracking even when only the language description is given.

\begin{figure}[t]
  \centering
 \vspace{-0pt}
   \includegraphics[width=1.0\linewidth]{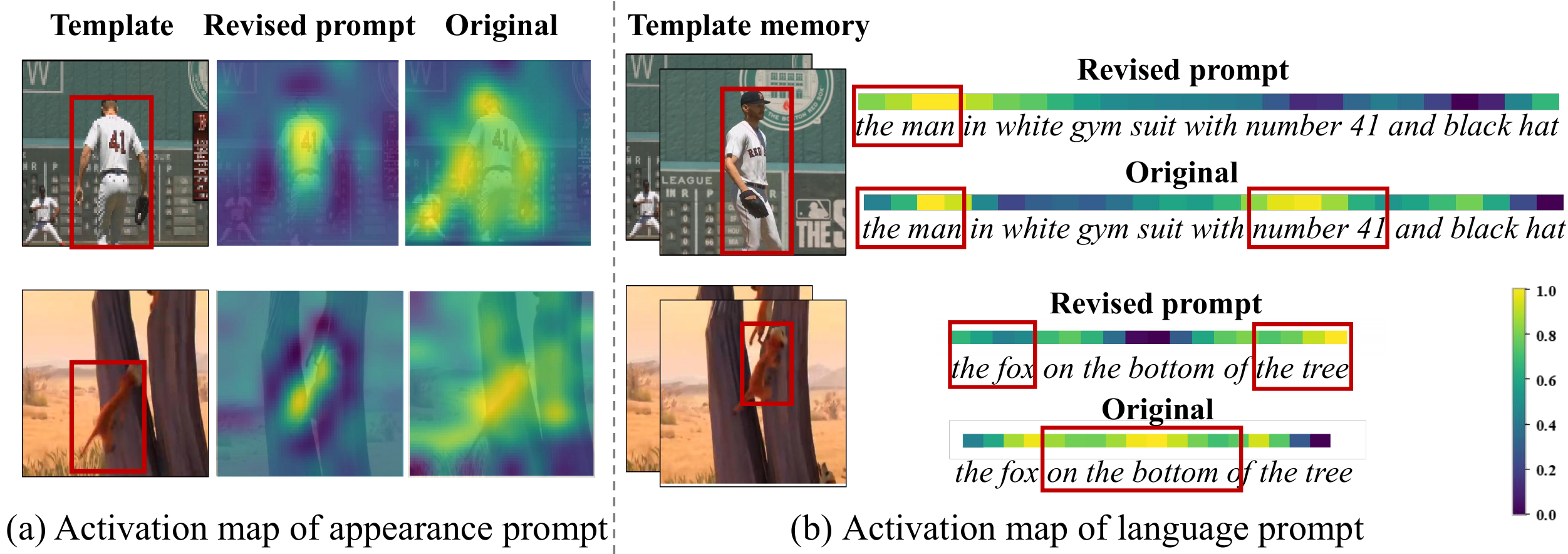}

   \caption{Visualization of the appearance and language prompts. 
   }
   \vspace{-10pt}
   \label{fig:vis_language}
\end{figure}

\section{Conclusion}
\quad 
In this paper, we have introduced a unified framework for natural language tracking that effectively leverages both visual and verbal references to improve target perception and discrimination.
We proposed the prompt modulation module to filter out the description in target references, thus forming accurate and context-aware visual and verbal cues. Besides, the target decoding module is designed to integrate multi-modal reference information to reason about the position of the target within the search image. Incorporating the target decoding network with precise target prompts greatly improves the discrimination of the tracker. Extensive experiments on the natural language tracking datasets and the visual grounding dataset demonstrate our proposed approach achieves competitive performance.


{
    \small
    \bibliographystyle{ieeenat_fullname}
    \normalem 
    \bibliography{main}

\begin{thebibliography}{43}
\providecommand{\natexlab}[1]{#1}
\providecommand{\url}[1]{\texttt{#1}}
\expandafter\ifx\csname urlstyle\endcsname\relax
  \providecommand{\doi}[1]{doi: #1}\else
  \providecommand{\doi}{doi: \begingroup \urlstyle{rm}\Url}\fi

\bibitem[Chen et~al.(2021)Chen, Yan, Zhu, Wang, Yang, and Lu]{TransT}
Xin Chen, Bin Yan, Jiawen Zhu, Dong Wang, Xiaoyun Yang, and Huchuan Lu.
\newblock Transformer tracking.
\newblock In \emph{IEEE Conf. Comput. Vis. Pattern Recog.}, pages 8126--8135, 2021.

\bibitem[Cui et~al.(2022)Cui, Guo, Shao, Wang, Shen, Zhang, and Chen]{cui2022joint}
Ying Cui, Dongyan Guo, Yanyan Shao, Zhenhua Wang, Chunhua Shen, Liyan Zhang, and Shengyong Chen.
\newblock Joint classification and regression for visual tracking with fully convolutional siamese networks.
\newblock \emph{Int. J. Comput. Vis.}, pages 1--17, 2022.

\bibitem[Deng et~al.(2021)Deng, Yang, Chen, Zhou, and Li]{deng2021transvg}
Jiajun Deng, Zhengyuan Yang, Tianlang Chen, Wengang Zhou, and Houqiang Li.
\newblock Transvg: End-to-end visual grounding with transformers.
\newblock In \emph{Int. Conf. Comput. Vis.}, pages 1769--1779, 2021.

\bibitem[Fan et~al.(2019)Fan, Lin, Yang, Chu, Deng, Yu, Bai, Xu, Liao, and Ling]{lasot}
Heng Fan, Liting Lin, Fan Yang, Peng Chu, Ge Deng, Sijia Yu, Hexin Bai, Yong Xu, Chunyuan Liao, and Haibin Ling.
\newblock Lasot: A high-quality benchmark for large-scale single object tracking.
\newblock In \emph{IEEE Conf. Comput. Vis. Pattern Recog.}, 2019.

\bibitem[Feng et~al.(2019)Feng, Ablavsky, Bai, and Sclaroff]{feng2019robust}
Qi Feng, Vitaly Ablavsky, Qinxun Bai, and Stan Sclaroff.
\newblock Robust visual object tracking with natural language region proposal network.
\newblock \emph{arXiv preprint arXiv:1912.02048}, 1\penalty0 (7):\penalty0 8, 2019.

\bibitem[Feng et~al.(2020)Feng, Ablavsky, Bai, Li, and Sclaroff]{2020wacv_feng}
Qi Feng, Vitaly Ablavsky, Qinxun Bai, Guorong Li, and Stan Sclaroff.
\newblock Real-time visual object tracking with natural language description.
\newblock In \emph{IEEE Conf. Appli. Comput. Vis.}, pages 700--709, 2020.

\bibitem[Feng et~al.(2021)Feng, Ablavsky, Bai, and Sclaroff]{2021cvpr_snlt}
Qi Feng, Vitaly Ablavsky, Qinxun Bai, and Stan Sclaroff.
\newblock Siamese natural language tracker: Tracking by natural language descriptions with siamese trackers.
\newblock In \emph{IEEE Conf. Comput. Vis. Pattern Recog.}, pages 5851--5860, 2021.

\bibitem[Filtenborg et~al.(2020)Filtenborg, Gavves, and Gupta]{2020arxiv_STL}
Maximilian Filtenborg, Efstratios Gavves, and Deepak Gupta.
\newblock Siamese tracking with lingual object constraints.
\newblock \emph{arXiv preprint arXiv:2011.11721}, 2020.

\bibitem[Guo et~al.(2021)Guo, Shao, Cui, Wang, Zhang, and Shen]{guo2021graph}
Dongyan Guo, Yanyan Shao, Ying Cui, Zhenhua Wang, Liyan Zhang, and Chunhua Shen.
\newblock Graph attention tracking.
\newblock In \emph{IEEE Conf. Comput. Vis. Pattern Recog.}, pages 9543--9552, 2021.

\bibitem[Guo et~al.(2022)Guo, Zhang, Fan, and Jing]{2022nips_guo}
Mingzhe Guo, Zhipeng Zhang, Heng Fan, and Liping Jing.
\newblock Divert more attention to vision-language tracking.
\newblock \emph{Adv. Neural Inform. Process. Syst.}, 35:\penalty0 4446--4460, 2022.

\bibitem[Huang et~al.(2021)Huang, Lian, Luo, and Gao]{LBYL}
Binbin Huang, Dongze Lian, Weixin Luo, and Shenghua Gao.
\newblock Look before you leap: Learning landmark features for one-stage visual grounding.
\newblock In \emph{IEEE Conf. Comput. Vis. Pattern Recog.}, pages 16888--16897, 2021.

\bibitem[Jang et~al.(2016)Jang, Gu, and Poole]{jang2016categorical}
Eric Jang, Shixiang Gu, and Ben Poole.
\newblock Categorical reparameterization with gumbel-softmax.
\newblock \emph{arXiv preprint arXiv:1611.01144}, 2016.

\bibitem[Krizhevsky et~al.(2017)Krizhevsky, Sutskever, and Hinton]{krizhevsky2017imagenet}
Alex Krizhevsky, Ilya Sutskever, and Geoffrey~E Hinton.
\newblock Imagenet classification with deep convolutional neural networks.
\newblock \emph{Comm. of the ACM}, 60\penalty0 (6):\penalty0 84--90, 2017.

\bibitem[Li et~al.(2018)Li, Yan, Wu, Zhu, and Hu]{2018cvpr_siamprn}
Bo Li, Junjie Yan, Wei Wu, Zheng Zhu, and Xiaolin Hu.
\newblock High performance visual tracking with siamese region proposal network.
\newblock In \emph{IEEE Conf. Comput. Vis. Pattern Recog.}, pages 8971--8980, 2018.

\bibitem[Li et~al.(2019)Li, Wu, Wang, Zhang, Xing, and Yan]{2019siamrpn++}
Bo Li, Wei Wu, Qiang Wang, Fangyi Zhang, Junliang Xing, and Junjie Yan.
\newblock Siamrpn++: Evolution of siamese visual tracking with very deep networks.
\newblock In \emph{IEEE Conf. Comput. Vis. Pattern Recog.}, pages 4282--4291, 2019.

\bibitem[Li et~al.(2023)Li, Huang, He, Wang, Lu, and Yang]{2023cvpr_cite}
Xin Li, Yuqing Huang, Zhenyu He, Yaowei Wang, Huchuan Lu, and Ming-Hsuan Yang.
\newblock Citetracker: Correlating image and text for visual tracking.
\newblock In \emph{IEEE Conf. Comput. Vis. Pattern Recog.}, pages 9974--9983, 2023.

\bibitem[Li et~al.(2022)Li, Yu, Cai, and Pan]{li2022cross}
Yihao Li, Jun Yu, Zhongpeng Cai, and Yuwen Pan.
\newblock Cross-modal target retrieval for tracking by natural language.
\newblock In \emph{IEEE Conf. Comput. Vis. Pattern Recog.}, pages 4931--4940, 2022.

\bibitem[Li et~al.(2017)Li, Tao, Gavves, Snoek, and Smeulders]{2017cvprLi}
Zhenyang Li, Ran Tao, Efstratios Gavves, Cees~GM Snoek, and Arnold~WM Smeulders.
\newblock Tracking by natural language specification.
\newblock In \emph{IEEE Conf. Comput. Vis. Pattern Recog.}, pages 6495--6503, 2017.

\bibitem[Lin et~al.(2022)Lin, Fan, Zhang, Xu, and Ling]{lin2022swintrack}
Liting Lin, Heng Fan, Zhipeng Zhang, Yong Xu, and Haibin Ling.
\newblock Swintrack: A simple and strong baseline for transformer tracking.
\newblock \emph{Adv. Neural Inform. Process. Syst.}, 35:\penalty0 16743--16754, 2022.

\bibitem[Liu et~al.(2019)Liu, Ott, Goyal, Du, Joshi, Chen, Levy, Lewis, Zettlemoyer, and Stoyanov]{liu2019roberta}
Yinhan Liu, Myle Ott, Naman Goyal, Jingfei Du, Mandar Joshi, Danqi Chen, Omer Levy, Mike Lewis, Luke Zettlemoyer, and Veselin Stoyanov.
\newblock Roberta: A robustly optimized bert pretraining approach.
\newblock \emph{arXiv preprint arXiv:1907.11692}, 2019.

\bibitem[Liu et~al.(2021)Liu, Lin, Cao, Hu, Wei, Zhang, Lin, and Guo]{liu2021swin}
Ze Liu, Yutong Lin, Yue Cao, Han Hu, Yixuan Wei, Zheng Zhang, Stephen Lin, and Baining Guo.
\newblock Swin transformer: Hierarchical vision transformer using shifted windows.
\newblock In \emph{Int. Conf. Comput. Vis.}, pages 10012--10022, 2021.

\bibitem[Maddison et~al.(2016)Maddison, Mnih, and Teh]{maddison2016concrete}
Chris~J Maddison, Andriy Mnih, and Yee~Whye Teh.
\newblock The concrete distribution: A continuous relaxation of discrete random variables.
\newblock \emph{arXiv preprint arXiv:1611.00712}, 2016.

\bibitem[Mao et~al.(2016)Mao, Huang, Toshev, Camburu, Yuille, and Murphy]{refcocog}
Junhua Mao, Jonathan Huang, Alexander Toshev, Oana Camburu, Alan~L Yuille, and Kevin Murphy.
\newblock Generation and comprehension of unambiguous object descriptions.
\newblock In \emph{IEEE Conf. Comput. Vis. Pattern Recog.}, pages 11--20, 2016.

\bibitem[Radford et~al.(2021)Radford, Kim, Hallacy, Ramesh, Goh, Agarwal, Sastry, Askell, Mishkin, Clark, et~al.]{2021_clip}
Alec Radford, Jong~Wook Kim, Chris Hallacy, Aditya Ramesh, Gabriel Goh, Sandhini Agarwal, Girish Sastry, Amanda Askell, Pamela Mishkin, Jack Clark, et~al.
\newblock Learning transferable visual models from natural language supervision.
\newblock In \emph{Int. Conf. Mach. Learn.}, pages 8748--8763. PMLR, 2021.

\bibitem[Real et~al.(2019)Real, Aggarwal, Huang, and Le]{real2019regularized}
Esteban Real, Alok Aggarwal, Yanping Huang, and Quoc~V Le.
\newblock Regularized evolution for image classifier architecture search.
\newblock In \emph{AAAI}, pages 4780--4789, 2019.

\bibitem[Rezatofighi et~al.(2019)Rezatofighi, Tsoi, Gwak, Sadeghian, Reid, and Savarese]{giou}
Hamid Rezatofighi, Nathan Tsoi, JunYoung Gwak, Amir Sadeghian, Ian Reid, and Silvio Savarese.
\newblock Generalized intersection over union: A metric and a loss for bounding box regression.
\newblock In \emph{IEEE Conf. Comput. Vis. Pattern Recog.}, pages 658--666, 2019.

\bibitem[Shao et~al.(2023)Shao, Ye, Luo, Zhang, and Chen]{shao2023intertracker}
Yanyan Shao, Qi Ye, Wenhan Luo, Kaihao Zhang, and Jiming Chen.
\newblock Intertracker: Discovering and tracking general objects interacting with hands in the wild.
\newblock In \emph{Int. Conf. on Intell. Robots and Systems}, pages 9079--9085. IEEE, 2023.

\bibitem[Wang et~al.(2021{\natexlab{a}})Wang, Zhou, Wang, and Li]{TrDimp}
Ning Wang, Wengang Zhou, Jie Wang, and Houqiang Li.
\newblock Transformer meets tracker: Exploiting temporal context for robust visual tracking.
\newblock In \emph{IEEE Conf. Comput. Vis. Pattern Recog.}, pages 1571--1580, 2021{\natexlab{a}}.

\bibitem[Wang et~al.(2018)Wang, Li, Yang, Zhang, Tang, and Luo]{wang2018describe}
Xiao Wang, Chenglong Li, Rui Yang, Tianzhu Zhang, Jin Tang, and Bin Luo.
\newblock Describe and attend to track: Learning natural language guided structural representation and visual attention for object tracking.
\newblock \emph{arXiv preprint arXiv:1811.10014}, 2018.

\bibitem[Wang et~al.(2021{\natexlab{b}})Wang, Shu, Zhang, Jiang, Wang, Tian, and Wu]{2021_tnl2k_Wang}
Xiao Wang, Xiujun Shu, Zhipeng Zhang, Bo Jiang, Yaowei Wang, Yonghong Tian, and Feng Wu.
\newblock Towards more flexible and accurate object tracking with natural language: Algorithms and benchmark.
\newblock In \emph{IEEE Conf. Comput. Vis. Pattern Recog.}, pages 13763--13773, 2021{\natexlab{b}}.

\bibitem[Wu et~al.(2015)Wu, Lim, and Yang]{wu2015object}
Yi Wu, Jongwoo Lim, and Ming-Hsuan Yang.
\newblock Object tracking benchmark.
\newblock \emph{IEEE Trans. Pattern Anal. Mach. Intell.}, 37\penalty0 (9):\penalty0 1834--1848, 2015.

\bibitem[Xu et~al.(2022)Xu, De~Mello, Liu, Byeon, Breuel, Kautz, and Wang]{xu2022groupvit}
Jiarui Xu, Shalini De~Mello, Sifei Liu, Wonmin Byeon, Thomas Breuel, Jan Kautz, and Xiaolong Wang.
\newblock Groupvit: Semantic segmentation emerges from text supervision.
\newblock In \emph{IEEE Conf. Comput. Vis. Pattern Recog.}, pages 18134--18144, 2022.

\bibitem[Yang et~al.(2022)Yang, Xu, Yuan, Liu, Li, and Hu]{VLTVG}
Li Yang, Yan Xu, Chunfeng Yuan, Wei Liu, Bing Li, and Weiming Hu.
\newblock Improving visual grounding with visual-linguistic verification and iterative reasoning.
\newblock In \emph{IEEE Conf. Comput. Vis. Pattern Recog.}, pages 9499--9508, 2022.

\bibitem[Yang et~al.(2020{\natexlab{a}})Yang, Chen, Wang, and Luo]{ReSC-Large}
Zhengyuan Yang, Tianlang Chen, Liwei Wang, and Jiebo Luo.
\newblock Improving one-stage visual grounding by recursive sub-query construction.
\newblock In \emph{Eur. Conf. Comput. Vis.}, pages 387--404, 2020{\natexlab{a}}.

\bibitem[Yang et~al.(2020{\natexlab{b}})Yang, Kumar, Chen, Su, and Luo]{2020tcsvt_yang}
Zhengyuan Yang, Tushar Kumar, Tianlang Chen, Jingsong Su, and Jiebo Luo.
\newblock Grounding-tracking-integration.
\newblock \emph{IEEE Trans. Circuit Syst. Video Technol.}, 31\penalty0 (9):\penalty0 3433--3443, 2020{\natexlab{b}}.

\bibitem[Ye et~al.(2022{\natexlab{a}})Ye, Chang, Ma, Shan, and Chen]{ostrack}
Botao Ye, Hong Chang, Bingpeng Ma, Shiguang Shan, and Xilin Chen.
\newblock Joint feature learning and relation modeling for tracking: A one-stream framework.
\newblock In \emph{Eur. Conf. Comput. Vis.}, pages 341--357, 2022{\natexlab{a}}.

\bibitem[Ye et~al.(2022{\natexlab{b}})Ye, Chang, Ma, Shan, and Chen]{ye2022joint}
Botao Ye, Hong Chang, Bingpeng Ma, Shiguang Shan, and Xilin Chen.
\newblock Joint feature learning and relation modeling for tracking: A one-stream framework.
\newblock In \emph{Eur. Conf. Comput. Vis.}, pages 341--357. Springer, 2022{\natexlab{b}}.

\bibitem[Zhang et~al.(2020)Zhang, Peng, Fu, Li, and Hu]{zhang2020ocean}
Zhipeng Zhang, Houwen Peng, Jianlong Fu, Bing Li, and Weiming Hu.
\newblock Ocean: Object-aware anchor-free tracking.
\newblock In \emph{Eur. Conf. Comput. Vis.}, pages 771--787, 2020.

\bibitem[Zhang et~al.(2021)Zhang, Liu, Wang, Li, and Hu]{zhang2021automatch}
Zhipeng Zhang, Yihao Liu, Xiao Wang, Bing Li, and Weiming Hu.
\newblock Learn to match: Automatic matching network design for visual tracking.
\newblock In \emph{Int. Conf. Comput. Vis.}, pages 13339--13348, 2021.

\bibitem[Zhao et~al.(2023)Zhao, Qi, and Wu]{zhao2023mind}
Chongyang Zhao, Yuankai Qi, and Qi Wu.
\newblock Mind the gap: Improving success rate of vision-and-language navigation by revisiting oracle success routes.
\newblock In \emph{ACM Int. Conf. Multimedia}, pages 4349--4358, 2023.

\bibitem[Zhou et~al.(2023)Zhou, Zhou, Mao, and He]{2023joint_zhou}
Li Zhou, Zikun Zhou, Kaige Mao, and Zhenyu He.
\newblock Joint visual grounding and tracking with natural language specification.
\newblock In \emph{IEEE Conf. Comput. Vis. Pattern Recog.}, pages 23151--23160, 2023.

\bibitem[Zhu et~al.(2020)Zhu, Su, Lu, Li, Wang, and Dai]{zhu2020deformable}
Xizhou Zhu, Weijie Su, Lewei Lu, Bin Li, Xiaogang Wang, and Jifeng Dai.
\newblock Deformable detr: Deformable transformers for end-to-end object detection.
\newblock \emph{arXiv preprint arXiv:2010.04159}, 2020.

\bibitem[Zoph and Le(2017)]{2017NAS}
Barret Zoph and Quoc~V Le.
\newblock Neural architecture search with reinforcement learning.
\newblock \emph{Int. Conf. Learn. Represent.}, 2017.

\end{thebibliography}
}


\end{document}